\documentclass{IEEEtran}
\usepackage{amsmath,amsfonts,amssymb}
\usepackage{algorithm,tabularx,setspace}
\usepackage[noend]{algpseudocode}
\usepackage{array}
\usepackage[caption=false,font=normalsize,labelfont=sf,textfont=sf]{subfig}
\usepackage{textcomp}
\usepackage{stfloats}
\usepackage{url}
\usepackage{verbatim}
\usepackage{graphicx}
\usepackage{cite}
\usepackage{subcaption}
\usepackage{algorithm}
\usepackage{algpseudocode}
\usepackage{mathtools}
\usepackage{amsthm}
\usepackage{bbm}
\usepackage{enumitem}
\usepackage{hyperref}
\usepackage{url}
\usepackage{colortbl}
\usepackage{makecell,multirow}       %
\usepackage{nicefrac}       %
\usepackage{thmtools}
\usepackage{xspace}
\usepackage{footnote, tablefootnote}
\usepackage{float}
\usepackage{epstopdf}
\usepackage{latexsym}
\usepackage{booktabs}
\floatstyle{plaintop}
\restylefloat{table}
\newtheorem{theorem}{Theorem}

\newtheorem{lemma}{Lemma}

\newcommand{\argmin}{\operatornamewithlimits{argmin}}

\begin{document}

\title{Bilevel Joint Unsupervised and Supervised Training for Automatic Speech Recognition}

\author{Xiaodong Cui$^{1}$, A F M Saif$^{2}$, Songtao Lu$^{1}$, Lisha Chen$^{2}$, Tianyi Chen$^{2}$, Brian Kingsbury$^{1}$,  George Saon$^{1}$ \\
$^{1}$IBM Research, IBM T. J. Watson Research Center, New York, USA \\
$^{2}$Rensselaer Polytechnic Institute, New York, USA}

\maketitle

\begin{abstract}
In this paper, we propose a bilevel joint unsupervised and supervised training (BL-JUST) framework for automatic speech recognition. Compared to the conventional pre-training and fine-tuning strategy which is a disconnected two-stage process, BL-JUST tries to optimize an acoustic model such that it simultaneously minimizes both the unsupervised and supervised loss functions. Because BL-JUST seeks matched local optima of both loss functions, acoustic representations learned by the acoustic model strike a good balance between being generic and task-specific. We solve the BL-JUST problem using penalty-based bilevel gradient descent and evaluate the trained deep neural network acoustic models on various datasets with a variety of architectures and loss functions. We show that BL-JUST can outperform the widely-used pre-training and fine-tuning strategy and some other popular semi-supervised techniques.
\end{abstract}

\begin{IEEEkeywords}
joint unsupervised and supervised training, bilevel optimization, automatic speech recognition, deep neural networks, semi-supervised training.
\end{IEEEkeywords}

\section{Introduction}
\label{sec:intro}

Deep learning~\cite{LeCun_DLNature} has made a great impact on automatic speech recognition (ASR) in the past decade. ASR systems with deep acoustic and language models have yielded unprecedented performance~\cite{Hinton_DNNSPM}. On some tasks, ASR can even achieve human parity \cite{Xiong_ASRParity,Saon_2017HumanParity}. The success of ASR with deep neural networks relies on large numbers of model parameters and large amounts of training data \cite{Parthasarathi_MillionhourASR,Parthasarathi_PetaAM}. Since labeled speech data is usually expensive to collect and transcribe \cite{deng2013new} while there are huge amounts of unlabeled speech data available in the field, a common way of training high-performing deep neural network acoustic models is a two-stage strategy. In this strategy, the acoustic models are first pre-trained (PT) using large amounts of unlabelled data under self-supervised training. Then the pre-trained models are fine-tuned (FT) using a small amount of labeled data via supervised training on downstream tasks. We will use PT+FT hereafter to refer to this two-stage pre-training followed by a fine-tuning approach. The PT+FT approach has been actively studied and yielded great performance in the speech community~\cite{baevski2019vq,baevski2020wav2vec,hsu2021hubert,Chiu_Bestrq}.

\begin{figure}[htb]
\centering
\centerline{\includegraphics[width=5.3cm, height=7.5cm]{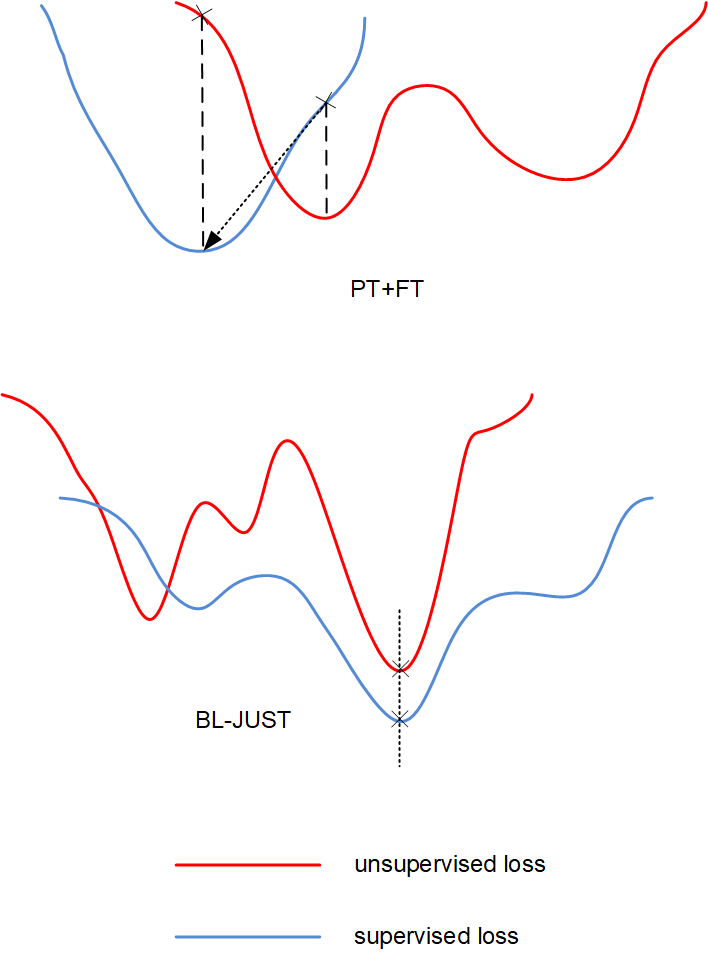}}
\caption{\small An illustration of the two-stage pre-training followed by fine-tuning (PT+FT) in the upper panel and bilevel joint unsupervised and supervised training (BL-JUST) in the lower panel.}\label{fig:ptft_just}
\end{figure}

One disadvantage of the PT+FT approach is that pre-training and fine-tuning are disconnected: the pre-training with unlabeled data is oblivious to the task-specific labeled data~\cite{rosenstein2005transfer}, and the downstream fine-tuning has limited control over the upstream self-supervised training.  This can be illustrated in the upper panel of Fig.~\ref{fig:ptft_just}. In the PT+FT strategy, the pre-training finds a local optimum in the unsupervised loss landscape from which the fine-tuning starts the optimization and eventually finds a local optimum in the supervised loss landscape. The only guidance provided by the unsupervised pre-training to the supervised fine-tuning is the initial model parameters. After that, the unsupervised loss is ignored in the following fine-tuning stage. This lack of interaction between the two loss functions may be problematic as the local optimum found in the fine-tuning stage may incur a high unsupervised loss, especially when the labeled data is limited, which may result in overfitting or even negative transfer~\cite{pan2009survey,wang2019characterizing}.

In this paper, we propose bilevel joint unsupervised and supervised training (BL-JUST) to overcome this limitation of the two-stage PT+FT.  The motivation of BL-JUST is to train a model that simultaneously minimizes both the unsupervised and supervised loss functions. In other words, we search for a matched pair of local optima of the two loss functions, which is illustrated in the lower panel of Fig.~\ref{fig:ptft_just}. Studies on the loss landscape of large deep neural networks suggest that these models possess large numbers of high-quality local minima~\cite{Choromanska2015}, and we therefore surmise that finding a local minimum with good performance for both loss functions is feasible. BL-JUST is carried out in a bilevel optimization framework which has seen increasing success in a wide variety of applications~\cite{liu2021investigating,crockett2022bilevel,chen2023learning,lu2023meta,franceschi2018bilevel,finn2017model}. Particularly, we treat the supervised training as the upper-level optimization problem. It is subject to a constraint of the lower-level optimization problem which is the unsupervised training. We solve the BL-JUST optimization using penalty-based bilevel gradient descent (PBGD)~\cite{shen2023penalty}. Since the unsupervised pre-training tends to capture generic acoustic representations while the supervised fine-tuning adapts to features oriented to specific downstream tasks, joint training could strike a good balance between the two. We extensively evaluate the acoustic models trained by the proposed BL-JUST on three diverse datasets using a variety of deep neural network architectures and loss functions. The results show that BL-JUST can outperform the widely-used PT+FT approach and some other popular semi-supervised techniques such as JUST \cite{bai2022joint}, pseudo labeling (PL)\cite{Xu_IPL} and alternating optimization (AO). \footnote{This work is a significant expansion of our previous work in \cite{Saif_bljust}.}.

The remainder of the paper is organized as follows. We formulate the BL-JUST problem in Section~\ref{sec:form}.  Section~\ref{sec:penbilevel} is devoted to the penalty-based bilevel optimization as a solver of BL-JUST. A pseudo-code implementation is given in Section~\ref{sec:impl}. Experimental results on LibriSpeech, Switchboard, and internal payload datasets are reported in Section~\ref{sec:exp}. We conclude the paper with a summary in Section~\ref{sec:sum}.

\section{Problem Formulation}
\label{sec:form}

\begin{figure}[htb]
\centering
\centerline{\includegraphics[height=6cm]{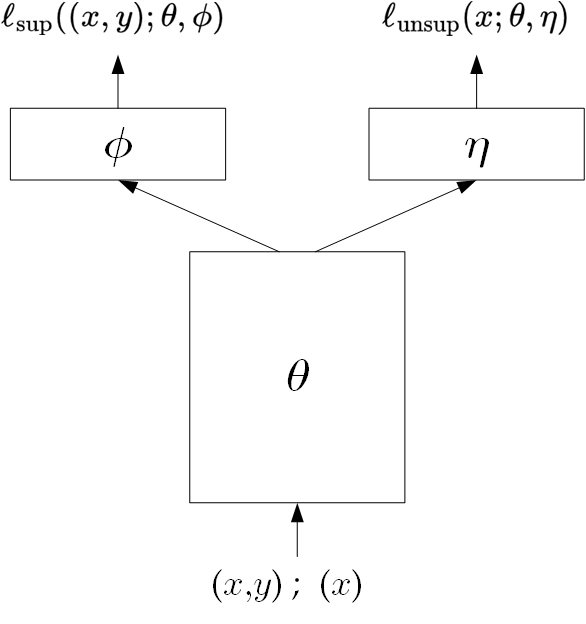}}
\caption{The network architecture for bilevel joint unsupervised and supervised training.}\label{fig:justconfig}
\end{figure}

Let $x$ denote the input speech feature sequences and $y$ the label sequences. Consider the network architecture in Fig.~\ref{fig:justconfig} where BL-JUST components are configured. Let $\theta$ be the parameters of the backbone network shared by both unsupervised and supervised training; and $\eta$ and $\phi$ be the parameters devoted to unsupervised and supervised training, respectively. We further denote $\ell_{\text{unsup}}(\cdot)$ as the unsupervised loss and $\ell_{\text{sup}}(\cdot)$ as the supervised loss.

Suppose $\mathcal{D}_{\text{sup}}$ is the labeled data and $\mathcal{D}_{\text{unsup}}$ is the unlabeled data. BL-JUST aims to solve the following bilevel optimization problem:
\begin{align}
\min_{\theta,\phi} & \sum_{(x,y)\in \mathcal{D}_{\text{sup}}} \ell_{\text{sup}}((x,y);\theta,\phi) \label{eqn:just_upper} \\
\mathrm{s.t.}  & \quad\theta \in \argmin_{\theta', \eta}\sum_{ x \in \mathcal{D}_{\text{unsup}}} \ell_{\text{unsup}}(x;\theta',\eta). \label{eqn:just_lower}
\end{align}

It can be seen that this is an optimization problem with hierarchy. Eq.~\ref{eqn:just_upper}, called the upper-level problem, minimizes the empirical risk of the supervised loss on the labeled data, and Eq.~\ref{eqn:just_lower}, called the lower-level problem, minimizes the empirical risk of the unsupervised loss on the unlabeled data. In the game-theoretic literature, the upper level is also referred to as the leader while the lower level is referred to as the follower. The leader optimizes its objective based on the follower's best response. Specialized to BL-JUST, the supervised training in the upper level (Eq.~\ref{eqn:just_upper}) optimizes the supervised empirical risk on the labeled data subject to the constraint of the optimal solution from the lower level (Eq.~\ref{eqn:just_lower}) unsupervised empirical risk on the unlabeled data.  Therefore, the supervised training is under the equilibrium constraints of the unsupervised loss. The optimal solution of BL-JUST will be shared (local) optima of both problems.

The idea of matched local optima for unsupervised and supervised loss functions has previously been investigated in~\cite{sutskever2015towards} through Output Distribution Matching (ODM), which measures the divergence between distributions of predictions and labels. The proposed BL-JUST seeks shared local optima through equilibrium constraints through a bilevel framework. Furthermore, ODM is only evaluated on a number of small datasets.

Choosing supervised learning as the upper-level problem and unsupervised learning as the lower-level problem is based on the following considerations. ASR acoustic modeling is ideally a supervised learning problem if every single speech signal has its ground-truth label. But in practice labeled data is both expensive and time-consuming to get and meanwhile unlabeled data is readily available. So typically unlabeled data is first used to learn acoustic representations which are then used to help the supervised learning with labeled data. This is especially true when there is only a small amount of labeled data and a large amount of unlabeled data. The acoustic representations learned from unsupervised learning may improve the robustness of supervised learning if it is appropriately leveraged. So we treat supervised learning as a leader problem (a.k.a. the upper-level problem) and unsupervised learning as a follower problem (a.k.a the lower-level problem). In addition, we require that the acoustic representations learned from unsupervised learning are of good quality such that some (local) optimality of the unsupervised loss has to be satisfied.

\section{Penalty-based Bilevel Gradient Descent}
\label{sec:penbilevel}

Although bilevel optimization has become increasingly popular to formulate and tackle various challenging problems in a broad variety of applications, it is difficult to solve in general due to its non-convex and non-differentiable nature. In recent years, various approaches based on implicit gradient and unrolled differentiation have been developed to solve bilevel optimization problems \cite{Zhang_Bilevel}. However, most of them are computationally demanding and not scale-friendly.  When dealing with complex constraints, value functions or penalty-based methods have proven to be effective in solving bilevel optimization problems \cite{shen2023penalty,Zhang_Bilevel,Lu_Bilevel,Kwon_penaltyBL}.  In this section, we use the PBGD algorithm~\cite{shen2023penalty} to solve the BL-JUST problem defined in Eqs.~\ref{eqn:just_upper} and \ref{eqn:just_lower}. PBGD provides a solution for bilevel optimization with general constraints and non-singleton lower-level optima.  The equivalence of solutions of bilevel optimization and its penalty-based reformulation is established based on the value function and KKT conditions in \cite{shen2023penalty}.

To keep the notations concise in the subsequent discussion, we define the generic upper- and lower-level objectives as
\begin{align}
f(\theta,\phi) & \triangleq \sum_{(x,y)\in \mathcal{D}_{\text{sup}}} \ell_{\text{sup}}((x,y);\theta,\phi) \\
g(\theta,\eta) & \triangleq \sum_{ x \in \mathcal{D}_{\text{unsup}}} \ell_{\text{unsup}}(x;\theta,\eta)
\end{align}
where $\theta \in \mathbb{R}^{d_{\theta}}$, $\phi \in \mathbb{R}^{d_{\phi}}$ and $\eta \in \mathbb{R}^{d_{\eta}}$.

Therefore, we can rewrite Eqs.~\ref{eqn:just_upper} and \ref{eqn:just_lower} as
\begin{equation}
\begin{aligned}
\min_{\theta,\phi} \quad & f(\theta,\phi)  \\
  \mathrm{s.t.} \quad & \theta \in  \argmin_{\theta',\eta}g(\theta',\eta).
\end{aligned}\label{eqn:bl}
\end{equation}
According to~\cite{shen2023penalty}, define the function value gap
\begin{align}
    p(\theta,\eta)  = g(\theta,\eta) - \min_{\theta,\eta}g(\theta,\eta)  \triangleq g(\theta,\eta) - v
\end{align}
where the constant value function $v$ is defined as
\begin{align}\label{eq:v_def}
 v  \triangleq \min_{\theta,\eta}g(\theta,\eta).
\end{align}
Thus we have
\begin{equation}
\begin{aligned}
\min_{\theta,\phi} \quad & f(\theta,\phi)   \\
\mathrm{s.t.} \quad & p(\theta,\eta) \leq 0.
\end{aligned}\label{eqn:blp}
\end{equation}
Note that Eq.~\ref{eqn:blp} is equivalent to Eq.~\ref{eqn:bl} but in the form of a function value gap.

To solve the bilevel problem in Eq.~\ref{eqn:blp}, PBGD uses the following penalty-based single-level reformulation
\begin{align}
\min_{\theta,\phi,\eta}~ F_{\gamma}(\theta,\phi,\eta) \triangleq & f(\theta,\phi) + \gamma p(\theta,\eta)  \label{eqn:psingle}
\end{align}
with a penalty factor $\gamma$. Note that $\gamma$ controls the strength of penalty on the function value gap function $p(\theta,\eta)$, not the original lower-level loss function $g(\theta,\eta)$. Then PBGD establishes the equivalence between this single-level reformulation in Eq.~\ref{eqn:psingle} with the original bilevel problem in Eq.~\ref{eqn:blp} based on Lemma~\ref{lemma:appequal} as long as certain assumptions are satisfied.

In fact, under the following assumptions that
\begin{enumerate}[label=(\alph*)]
\item the upper level loss $f(\theta,\phi)$ is $L$-Lipschitz continuous in $\phi$ given any $\theta$;
\item there exists $\mu\!>\!0$ such that
$g(\theta,\eta)$ satisfies the Polyak-{\L}ojasiewicz inequality
$\|\nabla g(\theta,\eta)\|^2 \geq \frac{1}{\mu}\big(g(\theta,\eta)-v\big)$ where $v$ is defined in Eq.~\ref{eq:v_def}; and
\item the gradient $\nabla F_\gamma(\theta,\phi,\eta)$ is $L_\gamma$-Lipschitz continuous;
\end{enumerate}
we have the following equivalence of two problems~\cite[Proposition 2]{shen2023penalty}:
\begin{lemma} \label{lemma:appequal}
Under the above assumptions, with a prescribed accuracy $\delta\!>\!0$, set $\gamma\!\geq\!L\sqrt{3\mu\delta^{-1}}$.
If $(\theta_\gamma,\phi_\gamma,\eta_\gamma)$ is a local/global solution of Eq.~\ref{eqn:psingle}, it is also a local/global solution of the following approximate problem of Eq.~\ref{eqn:blp}  with $\epsilon_\gamma\!\leq\!\delta$:
    \begin{equation}
    \begin{aligned}
        \min_{\theta,\phi} \quad & f(\theta,\phi)   \\
        \mathrm{s.t.} \quad & g(\theta,\eta) - \min_{\theta',\eta'}~g(\theta',\eta') \leq \epsilon_{\gamma}.
    \end{aligned}\label{eqn:blappr}
    \end{equation}
\end{lemma}
Lemma~\ref{lemma:appequal} tells us that for a sufficiently large $\gamma$ the global/local optima of the two problems are $\epsilon_{\gamma}$-approximately close. When $\gamma\!\rightarrow\!\infty$ the the global/local optima of the two problems will asymptotically match. In other words, the solution of Eq.~\ref{eqn:psingle} is also the solution of Eq.~\ref{eqn:blp} up to $\epsilon_{\gamma}$. As $\gamma$ increases, the tolerance $\epsilon_{\gamma}$ (i.e. gap to the optimum) decreases.

The general parameter update of $(\phi,\theta,\eta)$ in Eq.~\ref{eqn:psingle} under PBGD is given in \cite{shen2023penalty} as follows
\begin{multline}
(\phi_{k+1},\theta_{k+1},\eta_{k+1}) =  (\phi_{k},
\theta_{k},\eta_{k}) - \alpha [ \nabla f(\theta_{k},\phi_{k} )   \\
 +\gamma (\nabla g(\theta_{k},\eta_{k}) -  \nabla v ) ].
\label{eqn:pbgd}\end{multline}
Specialized to our BL-JUST configuration in Fig.~\ref{fig:justconfig}, $f(\theta,\phi)$ is independent of $\eta$ while $g( \theta,\eta)$ is independent of $\phi$. Furthermore, both $g(\theta,\eta)$ and $v$ are independent of $\phi$. Under these conditions, we have

\begin{align}
   \nabla v  =  \mathbf{0},  \ \ \ \ \mathbf{0} \in \mathbb{R}^{d_{\phi}+d_{\theta}+d_{\eta}} \label{eqn:pbgd_g_allzero}
\end{align}
and Eq.~\ref{eqn:pbgd} can be simplified to
\begin{multline}
(\phi_{k+1},\theta_{k+1},\eta_{k+1}) =  (\phi_{k},
\theta_{k},\eta_{k}) \\  - \alpha [ \nabla f(\phi_{k},\theta_{k}) + \gamma \nabla g(\theta_{k},\eta_{k}) ].  \label{eqn:pbgd_just}
\end{multline}
Eq.~\ref{eqn:pbgd_just} is used for updating parameters in Fig.~\ref{fig:justconfig}.

The convergence of the PBGD given in Eq.~\ref{eqn:pbgd_just} can be established by~\cite[Theorem 3]{shen2023penalty} under the above assumptions.
\begin{theorem}[Convergence rate of BL-JUST]\label{theorem:convergence}
Select an accuracy $\delta$ and $\beta\in (0, L_\gamma^{-1}]$,~with $\gamma$  chosen by Lemma~\ref{lemma:appequal}. Let $C\!=\!\inf_{(\phi,\theta)}f(\phi,\theta)$. Then it holds that
\begin{equation}
\frac{1}{N}\sum_{i=1}^N \|\nabla F_\gamma(\phi_i,\theta_i,\eta_i)\|^2 \leq \frac{18\big(F_\gamma(\phi_1,\theta_1)-C\big)}{\beta N}+\frac{10 L^2 L_\gamma^2}{N} \nonumber
\end{equation}
where $N$ is the number of iterations.
\end{theorem}
This theorem suggests the iteration complexity to achieve $\epsilon$-stationary point is $\mathcal{O}(L_\gamma\epsilon^{-1})$, which matches the complexity of the gradient descent-based supervised training method~\cite{bottou2018optimization}.

\section{Implementation}
\label{sec:impl}

In bilevel optimization, the feasible area of the upper-level problem is determined in part by the solution set of the lower-level problem. A penalty schedule is used in \cite{shen2023penalty} where the penalty factor starts from 0 and monotonically increases to its maximum value over epochs. Such a penalty schedule initializes the optimization of the upper-level problem without any constraint from the lower level and gradually shrinks the tolerance of the lower-level problem to a sufficiently small value during training. Algorithm~\ref{alg:bl_just} gives the implementation of BL-JUST in practice. In this implementation, we also conduct in each epoch a self-supervised exploration by approximately solving the lower-level unsupervised problem using SGD. The self-supervised exploration brings the optimizer to the proximity of a local optimum of the unsupervised loss. In the next step, we carry out joint unsupervised and supervised training based on PBGD according to Eq.~\ref{eqn:pbgd_just}. This step involves double sampling of batches from both labeled and unlabeled data under the supervised and unsupervised loss functions respectively. The penalty factor $\gamma$ is linearly increased in each epoch from 0 to a pre-defined maximum value.  In the end, a supervised fine-tuning with a small learning rate is conducted to further search the local optimum within the resulting $\epsilon_{\gamma}$ tolerance.

\begin{algorithm}[H]
\caption{Bilevel Joint Unsupervised and Supervised Training (BL-JUST)}
\setstretch{1.2}
\begin{algorithmic}
\State  Input:  labeled data $(x,y)$, unlabeled data $x$
\State  $\rho \leftarrow$ learning rate for self-supervised exploration
\State  $\alpha\leftarrow$ learning rate for bilevel gradient descent
\State  $\tau \leftarrow$ learning rate for supervised fine-tuning
\State  $\gamma_{m} \leftarrow$  maximum penalty factor
\State  $K \leftarrow$ number of epochs
\State  $N_{1} \leftarrow$ number of iterations in unsupervised training
\State  $N_{2} \leftarrow$ number of iterations in joint unsupervised and supervised training;
\State  $N_{3} \leftarrow$ number of iterations for supervised fine-tuning
\State  $L_{\text{unsup}} \leftarrow$ lower level unsupervised empirical risk $g(\theta, \eta)$
\State  $L_{\text{sup}} \leftarrow$ upper level supervised empirical risk $f(\theta, \phi)$
\vspace{0.2cm}
\For{$k = 1 : K$}
    \smallskip
    \State \textbf{\% self-supervised exploring}
    \For{$i = 1 : N_{1}$}
        \State $(\theta^{k}_{i+1},\eta^{k}_{i+1}) = (\theta^{k}_i,\eta^{k}_i) - \rho \nabla_{\theta,\eta} L_{\text{unsup}}(\theta^{k}_i,\eta^{k}_i)$
    \EndFor
    \smallskip
    \State \textbf{\% joint unsupervised and supervised training}
    \State  $\gamma_{k} = (k-1)\frac{\gamma_{m}}{K}$
    \State Use $\theta^{k}_{N_{1}+1}$ and $\eta^{k}_{N_{1}+1}$ as the starting point
    \For{$j = 1 : N_{2}$}
        \State Update $\theta^{k}_{j+1}$:
        \State \ $\theta^{k}_{j+1}\!=\!\theta^{k}_j - \alpha \nabla_\theta L_{\text{sup}}(\theta^{k}_j,\phi^{k}_j)
                                -\alpha\gamma_{k}\nabla_\theta L_{\text{unsup}}(\theta^{k}_j,\eta^{k}_j)$
        \State Update $\phi^{k}_{j+1}$:
        \State \ $\phi^{k}_{j+1} = \phi^{k}_j - \alpha \nabla_{\phi} L_{\text{sup}}(\theta^{k}_j,\phi^{k}_j)$
        \State Update $\eta^{k}_{j+1}$:
        \State \ $\eta^{k}_{j+1} = \eta^{k}_j - \alpha\gamma\nabla_{\eta} L_{\text{unsup}}(\theta^{k}_j,\eta^{k}_j)$
        \EndFor
\EndFor
\smallskip
\State \textbf{\% supervised fine-tuning}
\State Use $\theta^{K}_{N_{2}+1}$ and $\phi^{K}_{N_{2}+1}$ as the starting point
\For{$t = 1 : N_{3}$}
    \State  $(\theta_{t+1},\phi_{t+1}) = ( \theta_{t}, \phi_{t}) - \tau \nabla_{\theta,\phi} L_{\text{sup}}(\theta_t,\phi_t)$
\EndFor
\end{algorithmic}
\label{alg:bl_just}
\end{algorithm}

Joint unsupervised and supervised training has been applied in acoustic modeling. For instance, it is used in multilingual ASR in \cite{bai2022joint} and \cite{Zhang_USM} where the loss function is a weighted sum of unsupervised loss and supervised loss. In particular, the architectural design of the network in \cite{bai2022joint} in which unsupervised learning and supervised learning modules are stacked sequentially gives rise to a similar configuration as Fig.\ref{fig:justconfig}.  The major difference is that BL-JUST has the unsupervised exploration and, more importantly, it uses the local optimum of the unsupervised loss as a constraint to the optimization of supervised loss. The latter is accomplished through a monotonically increasing penalty schedule. Therefore, JUST in \cite{bai2022joint} is essentially a BL-JUST with a constant penalty factor and without unsupervised exploration.

\section{Experiments}
\label{sec:exp}

This section presents experimental results on three datasets including LibriSpeech, Switchboard, and Payload.  The first two are public speech datasets and the third is an in-house industrial speech dataset. These three datasets are chosen to provide broad coverage of size, speaking style, sampling rate, and domains. We evaluate the BL-JUST algorithm on these three diverse datasets with various unsupervised and supervised loss functions using various neural network architectures. Other than comparing with PT+FT, we also compare the performance of the proposed BL-JUST with other existing joint training or semi-supervised training approaches in the literature.

\subsection{Comparing with PT+FT}
\label{sec:comp_pt_ft}

\subsubsection{LibriSpeech} \hspace{0pt}
\label{sec:libri}

\textbf{Dataset.} \ LibriSpeech~\cite{panayotov2015librispeech} is a large English corpus consisting of 960 hours of read speech sampled at 16kHz. The training sets consist of 100 hours and 360 hours of clean data with transcribed text labels, called train-clean-100 and train-clean-360, respectively. Additionally, there is a subset of training data known as train-other-500, consisting of 500 hours of noisy data. There are two test sets labeled as test-clean and test-other.

\textbf{Model.} \ The acoustic model is a Conformer~\cite{Gulati_conformer} consisting of 10 Conformer blocks.  The input to the model is 80-dimensional log-Mel spectrogram features. Each Conformer block has 612 hidden units with 12 attention heads. Each head has a dimensionality of 51. The convolution kernel size used in each Conformer block is 31.  The final output layer has 1000 softmax units which correspond to the output classes. We use SentencePiece~\cite{kudo2018sentencepiece}, a language-independent subword tokenizer and detokenizer, to generate the 1000 output units. There are 100M parameters in this Conformer acoustic model.

\textbf{Regularization.} \ We use SpecAug~\cite{Park_SpecAug} and additionally employ dropout with a rate of 0.1 in each residual unit of the Conformer.

\textbf{Training strategies.}
We consider three acoustic models trained using strategies -- supervised baselines, PT+FT, and BL-JUST. We also train models under three conditions using varying amounts of unlabeled and labeled data:
\begin{itemize}
\item[a)]  using the same 100 hours of data (train-clean-100) as the unlabeled and labeled datasets.
\item[b)]  using the same 960 hours of data (train-clean-100 + train-clean-360 + train-other-500) as the unlabeled and labeled datasets.
\item[c)]  using the 860 hours of data (train-clean-360 + train-other-500) as the unlabeled dataset and 100 hours of data (train-clean-100) as the labeled dataset.
\end{itemize}

The supervised models are trained in each condition using the respective supervised datasets as baselines. For PT+FT, unsupervised pre-training is carried out using the Contrastive Predictive Coding (CPC) loss~\cite{oord2018representation} on the unlabeled datasets. The fine-tuning is carried out using the Connectionist Temporal Classification (CTC) loss \cite{graves2006connectionist} on the labeled datasets. For BL-JUST, the upper-level problem is supervised training with the CTC loss on the labeled datasets, and the lower-level problem is unsupervised training with the CPC loss on the unlabeled datasets, respectively. Details of the training recipes can be found in Appendix~\ref{app:libri}.

\textbf{Decoding.} \  A beam search decoder is used for decoding without external language models (LMs).

\textbf{Performance.} \  Table~\ref{tab:ls} presents the WERs of the baseline supervised model, as well as acoustic models trained using PT+FT and BL-JUST under different data conditions.  It can be observed from the table that the ASR performance in terms of WER exhibits similar trends in supervised baseline, PT+FT, and BL-JUST on both test sets in three data conditions. With the help of unsupervised training using the unlabeled data, both PT+FT and BL-JUST can outperform the supervised baseline.  In addition, BL-JUST yields superior performance over PT+FT in all cases.  Specifically, the WERs are 4.9\% vs.\ 5.8\%, 1.8\% vs.\ 2.0\%, 4.1\% vs.\ 5.1\% on test-clean and 12.2\% vs.\ 14.0\%, 3.4\% vs.\ 3.9\%, 11.3\% vs.\ 13.2\% on test-other in 100h/100h, 960h/960h and 100h/860h cases, respectively. Furthermore, with more unlabeled data, the performance of both PT+FT and BL-JUST improves, which can be seen by comparing the WERs on the two respective test sets under the 100h/100h and 100h/860h cases in the table.

\begin{table}[tbh]
\centering
\begin{tabular}{ l c  r  r  r  } \hline
    model                 &   hours          &     size      &      test-clean    &   test-other        \\
                          &   (L/U)          &               &                    &                     \\ \hline\hline
sup. baseline             &   100/--         &     100M      &        6.3         &     14.7            \\
PT+FT                     &   100/100        &     100M      &        5.8         &     14.0            \\
BL-JUST                   &   100/100        &     100M      &        4.9         &     12.2            \\  \hline
sup. baseline             &   960/--         &     100M      &        2.2         &      4.6            \\
PT+FT                     &   960/960        &     100M      &        2.0         &      3.9            \\
BL-JUST                   &   960/960        &     100M      &        1.8         &      3.4            \\  \hline
sup. baseline             &   100/--         &     100M      &        6.3         &     14.7            \\
PT+FT                     &   100/860        &     100M      &        5.1         &     13.2             \\
BL-JUST                   &   100/860        &     100M      &        4.1         &     11.3             \\ \hline
\end{tabular}
\caption{WERs on LibriSpeech for supervised CTC baselines and CTC acoustic models trained using PT+FT and BL-JUST. The models are evaluated under three conditions with different amounts of unlabeled and labeled data.}\label{tab:ls}
\end{table}

\subsubsection{Switchboard} \hspace{0pt}
\label{sec:swb}

\textbf{Dataset.} \ The Switchboard dataset consists of 2000 hours of English conversational speech sampled at 8kHz, which will be referred to as SWB2000. A 300-hour set is often selected and commonly used as a moderate dataset for experiments, which will be referred to as SWB300.  Speed and tempo perturbation is conducted offline on the 2000-hour speech, which gives rise to 4 replicas of the original speech data. Therefore, the total amount of training data is around 10,000 hours for SWB2000 and  1,500 hours for SWB300 after speed and tempo perturbation.  We measure WERs on the NIST Hub5 2000 (Switchboard and CallHome) test sets.

\textbf{Model.} \ The acoustic model is an RNN-Transducer (RNNT)~\cite{Graves_RNNT,Graves_RNNASR,Saon_RNNT,Li_RNNT}. The transcription network of the RNNT is a Conformer, the input to which is 40-dimensional log-Mel spectrogram features and their first and second-order derivatives. Features of every two adjacent frames are concatenated which gives 240-dimensional input vectors. Each conformer block has 512 hidden units and 8 attention heads of 64 dimensions. The convolution kernel size is 31. The prediction network is a single-layer uni-directional LSTM  with 1024 cells. The outputs of the transcription network and the prediction network are projected down to a 256-dimensional latent space in the joint network. The softmax layer contains 46 output units which correspond to 45 characters and the null symbol.

\textbf{Regularization.} \ Other than the offline speed and tempo perturbation, two additional data augmentation techniques, namely, noise injection~\cite{Saon_mixup} and SpecAug~\cite{Park_SpecAug} are conducted on the fly. The training is also regularized by dropout with a dropout rate of 0.1 for the conformer, 0.25 for the LSTM, and 0.05 for the embedding. In addition, DropConnect~\cite{Wan_dropconnect} is applied with a rate of 0.25 to the LSTM.

\textbf{Training strategies.} \  We treat SWB2000 as the unlabeled dataset and SWB300 as the labeled dataset. We consider three acoustic models trained using different training strategies -- supervised baselines, PT+FT, and BL-JUST. The supervised models are trained using SWB300 and SWB2000, respectively, as baselines. For PT+FT, the unsupervised pre-training on SWB2000 is carried out using BEST-RQ~\cite{Chiu_Bestrq}. The loss function is the cross-entropy on the masked frames. The fine-tuning on SWB300 is carried out under the RNNT loss.  For BL-JUST, the upper-level problem is SWB300 supervised training with the RNNT loss and the lower-level problem is SWB2000 unsupervised BEST-RQ training with the masked cross-entropy loss.  Details of the training recipes can be found in Appendix~\ref{app:swb}.

\textbf{Decoding.} \  Inference uses alignment-length synchronous decoding~\cite{Saon_RNNTdecoding}, which only allows hypotheses with the same alignment length in the beam for the beam search. No external LMs are used.

\textbf{Performance.} \  Table~\ref{tab:swb} presents the WERs of supervised baselines as well as acoustic models trained under PT+FT and BL-JUST. The supervised baselines on SWB300 and SWB2000 are RNNT acoustic models trained using 300-hour and 2000-hour labeled data (with data augmentation), respectively. The SWB300 transcription network consists of 6 conformer blocks while the SWB2000 transcription network consists of 10 conformer blocks.  The WERs are 11.2\% for SWB300 and 7.0\% for SWB2000. In PT+FT, the conformer has 10 blocks. We first carry out BEST-RQ training using labels created by aligning log-Mel spectrogram features against the random codebook, which gives 10.9\% WER, denoted by PT+FT/logmel. Inspired by HuBERT~\cite{hsu2021hubert}, we refine the frame labels by re-aligning intermediate features (output of the 7th block) of the trained conformer against the random codebook. This process is repeated twice, denoted by PT+FT/halign and PT+FT/halign$^{2}$ respectively in the table. It can be seen that PT+FT reduces WER over the SWB300 supervised baseline with the help of pre-training on the 2000-hour unlabeled dataset ($11.2\%\!\rightarrow\!10.9\%$). Label re-alignment using intermediate features can further reduce the WER ($10.9\%\!\rightarrow\!9.6\%$). In BL-JUST, the conformer is composed of 10 conformer blocks. We use the re-aligned frame labels in the BEST-RQ in the BL-JUST lower-level problem. This gives 8.2\% WER on average which is 1.4\% absolute better than that of PT+FT. If we use a larger model with 15 conformer blocks, BL-JUST can achieve a WER of 8.1\%.

\begin{table}[tbh]
\centering
\begin{tabular}{ l c  r  c  c  r } \hline
    model                 &   hours          &     size      &        SWB   &    CH    &   Avg       \\
                          &   (L/U)          &               &              &          &             \\ \hline\hline
sup. baseline             &   300/--         &     45M       &        7.4   &   15.0   &   11.2      \\
sup. baseline             &  2000/--         &     74M       &        5.5   &    8.5   &    7.0      \\ \hline
PT+FT/logmel              &   300/2000       &     74M       &        7.2   &   14.6   &   10.9      \\
PT+FT/halign              &   300/2000       &     74M       &        6.4   &   13.3   &    9.8      \\
PT+FT/halign$^{2}$        &   300/2000       &     74M       &        6.4   &   12.8   &    9.6      \\ \hline
BL-JUST                   &   300/2000       &     74M       &        5.8   &   10.6   &    8.2      \\
BL-JUST                   &   300/2000       &    109M       &        5.8   &   10.5   &    8.1      \\ \hline
\end{tabular}
\caption{NIST Hub5 2000 Switchboard and CallHome test set WERs for supervised RNNT baselines and RNNT acoustic models trained under PT+FT and BL-JUST. SWB2000 is used as unlabeled data while SWB300 is used as labeled data. }\label{tab:swb}
\end{table}

\subsubsection{Payload} \hspace{0pt}
\label{sec:payload}

\textbf{Dataset.} \ The in-house payload dataset consists of 90,000 hours of unlabeled and 10,000 hours of labeled English spontaneous speech recordings which were collected from real-world ASR services. The sampling rate is 8kHz. WERs are measured on 6 test sets varying in duration from 1.4 to 7.3 hours with an average of 3.7 hours. They represent a good coverage of application domains in model deployment.  We report the average WERs across the 6 test sets (S$_{1}$ to S$_{6}$).

\textbf{Model.} \ The acoustic model is a CTC conformer model. The input is 40-dimensional log-Mel spectrogram features and their first and second-order derivatives. Features of every two adjacent frames are concatenated which gives 240-dimensional input vectors. Each conformer block has 768 hidden units and 8 attention heads of 96 dimensions. The convolution kernel size is 31. The softmax layer contains 43 output units which correspond to 42 characters and the null symbol.

\textbf{Regularization.} \ Speed and tempo perturbation is conducted offline for data augmentation. Noise injection and SpecAug are conducted on the fly. The dropout rate for the conformer is 0.1.

\textbf{Training strategies.} \  We consider three acoustic models trained using different training strategies -- supervised baselines, PT+FT, and BL-JUST. The supervised models are trained using 10,000 hours of labeled data as the baseline. For PT+FT, the unsupervised pre-training on 90,000 hours of unlabeled data is carried out using BEST-RQ. The loss function is the cross-entropy on the masked frames. The fine-tuning on 10,000 hours of labeled data is carried out using the CTC loss. For BL-JUST, the upper-level problem is supervised training with the CTC loss using 10,000 hours of labeled data, and the lower-level problem is unsupervised BEST-RQ training with the masked cross-entropy loss using 90,000 hours of unlabeled data. Details of the training recipes can be found in Appendix~\ref{app:payload}.

\textbf{Decoding.} \  A beam search decoder is used for decoding without external LMs.

\textbf{Performance.} \ Table~\ref{tab:payload} compares the average WER across six test sets for the supervised baseline, PT+FT, and BL-JUST. It can be seen that in this case, when pre-training on 90,000 hours of unlabeled data and then fine-tuning on 10,000 hours of labeled data, PT+FT only gives a slight improvement from 16.0\% to 15.7\% over the supervised baseline trained on the same 10,000 hours of labeled data. Meanwhile, BL-JUST can improve the WER to 14.8\%, which outperforms both the supervised baseline and PT+FT.

\begin{table*}[tbh]
\centering
\begin{tabular}{ l  c  r  c c c c c c   c } \hline
    model             &    hours       &     size      &   S$_{1}$  &  S$_{2}$  &  S$_{3}$ &  S$_{4}$  &  S$_{5}$  &   S$_{6}$    &    avg.  \\
                      &     (L/U)       &        &    &     &     &     &     &       &          \\ \hline\hline
sup. baseline         &    10K/--      &     74M       &  6.2  &  21.3 &  14.6 &  15.9 &  12.2 &  25.8   &  16.0    \\
PT+FT                 &    10K/90K     &     83M       &  5.7  &  19.5 &  13.5 &  16.7 &  12.3 &  26.2   &  15.7    \\
BL-JUST               &    10K/90K     &     83M       &  5.3  &  19.1 &  12.8 &  15.2 &  12.1 &  24.3   &  14.8    \\ \hline
\end{tabular}
\caption{Averaged WERs on supervised CTC baselines and CTC acoustic models trained under PT+FT and BL-JUST across 6 test sets. In PT+FT and BL-JUST, there are 90,000 hours of unlabeled data and 10,000 hours of labeled data. }\label{tab:payload}
\end{table*}

\subsection{Ablation study}

In Algorithm \ref{alg:bl_just}, we conduct self-supervised exploration along with joint unsupervised and supervised training in each epoch. In the end, we also apply fine-tuning (FT) with a small learning rate. To investigate their impact, we carry out an ablation study by removing the self-supervised exploration step and the final FT step, respectively, using the 100h/860h data split of Librispeech. The resulting WERs are presented in Table~\ref{tab:abalation}. As can be observed from the table, removing fine-tuning in the last step results in 0.5\% and 0.3\% absolute degradation in WER on test-clean and test-other, respectively. Meanwhile, removing the self-supervised exploring step incurs 1.2\% and 1.6\% absolute degradation in WER on test-clean and test-other, respectively, which is more severe than that of fine-tuning.  Therefore, both steps are helpful in improving the performance of BL-JUST but the self-supervised exploration helps more. If we remove both steps, the WERs will significantly degrade to 5.9\% and 13.1\% compared to the original BL-JUST WER 4.1\% and 11.3\% on test-clean and test-other, respectively.
\begin{table}[tbh]
\centering
\begin{tabular}{ l  r  r  r  } \hline
    Configurations                   &      test-clean    &   test-other        \\ \hline\hline
    BL-JUST                          &        4.1         &     11.3            \\
    \ \  -- fine-tuning              &        4.6         &     11.6            \\
    \ \  -- self-supervised explr.   &        5.3         &     12.9            \\
    \ \ \ \ \ \  -- fine-tuning      &        5.9         &     13.1         \\ \hline
\end{tabular}
\caption{WERs of BL-JUST on LibriSpeech by removing self-supervised exploration and fine-tuning.  The 100h/860h labeled/unlabeled data split and the acoustic model of 100M parameters are used in this ablation study. }\label{tab:abalation}
\end{table}

\subsection{Comparing to other approaches}
\label{sec:comp}

In this section, we compare the proposed BL-JUST with a number of approaches to joint unsupervised and supervised training and semi-supervised training in the literature. Specifically, we compare with JUST \cite{bai2022joint}, pseudo-labeling (PL), and alternating optimization (AO).

\textbf{Comparing to JUST} \ As discussed, JUST in \cite{bai2022joint} is essentially a BL-JUST with a constant penalty factor and without unsupervised exploration. In Table~\ref{tab:just} we investigate the performance of BL-JUST under various schedules of $\gamma$ and compare it to the case when a constant $\gamma$ is used. Then we compare with JUST which is trained using a constant $\gamma$ without unsupervised exploration. The experiments are again carried out on LibriSpeech using 860 hours of unlabeled data and 100 hours of labeled data (L/U: 100/860 in Table~\ref{tab:ls}). The first block in the table presents the WERs on test-clean and test-other using $\gamma$ with various rates and maximum values. The best WERs are obtained using a rate of 0.0002 with a final maximum value of 0.2. The second block presents WERs when BL-JUST uses a constant $\gamma$ and JUST. In both cases, the training starts with randomized weights without unsupervised pre-training. It can be observed that using a constant penalty in BL-JUST leads to higher WERs.  Similarly, JUST
also gives higher WERs. The third block repeats the same setting as that of the second block except that both training is initialized with a pre-trained model after the unsupervised training on 860 hours of unlabeled data. It can be observed that with pre-training both BL-JUST with a constant penalty and JUST can significantly improve WERs over those without pre-training. However, even with pre-training
JUST still can not outperform BL-JUST. The results suggest that simply applying a constant penalty is not optimal to solving this bilevel optimization problem. In addition, in order to yield good performance when a constant penalty is used appropriate model initialization strategies have to be applied.

\begin{table}[tbh]
\centering
\begin{tabular}{c  c  c  c  c  c  c  } \hline
method       &           PT           &    init       &    inc.          &    max         &   test        &   test    \\
                         &            &    value      &    rate          &    value       &   clean       &   other   \\ \hline\hline
\multirow{4}{*}{BL-JUST} &  \multirow{4}{*}{N}        &     0.0      &   0.0010         &    0.10        &    6.7        &    14.8   \\
                         &            &     0.0       &   0.0020         &  0.20 &  4.1 &  11.3   \\
                         &            &     0.0       &   0.0025         &    0.25        &    4.7        &    12.1      \\
                         &            &     0.0       &   0.0030         &    0.30        &    5.2        &    13.6      \\ \hline\hline
\multirow{3}{*}{\shortstack{BL-JUST\\(constant $\gamma$)}} &  \multirow{3}{*}{N}        &    0.10      &      0.0           &    0.10                             &    8.6        &    16.5 \\
                         &            & 0.20 &  0.0     &  0.20 &  8.4 &  16.1   \\
                         &            simi&    0.30  &      0.0           &    0.30        & 8.9          &    16.7    \\ \hline
\multirow{4}{*}{JUST}    &  \multirow{4}{*}{N} & 0.03 & 0.0 & 0.03 &  6.7 & 14.2 \\ &   &   0.05      &       0.0             &       0.05         &      6.5       &     14.8  \\
                         &            & 0.10 &  0.0     &  0.10 & 8.8  &  17.1    \\
                         &            &    0.20  &      0.0           &    0.20        &     9.2       &    17.3    \\ \hline\hline
\multirow{3}{*}{\shortstack{BL-JUST\\(constant $\gamma$)}} &  \multirow{3}{*}{Y}        &    0.10      &     0.0              & 0.10               &     5.8          &     15.3   \\
                         &            & 0.20 &  0.0     &  0.20 &  5.0 &  12.9   \\
                         &            &    0.30      &              0.0  &  0.30         & 5.4          & 13.7       \\ \hline
\multirow{3}{*}{JUST}    &  \multirow{3}{*}{Y}         &    0.05     &       0.0             &        0.05        &   6.1            &   15.4    \\
                         &            & 0.10 &   0.0   & 0.10   & 5.5  & 14.9    \\
                         &            &    0.20     &  0.0      &  0.20  &   5.7  &   13.0 \\\hline\end{tabular}
\caption{WERs of BL-JUST under various penalty schedules and it comparison to BL-JUST with a constant penalty and JUST in \cite{bai2022joint} with and without pre-training.  The 100h/860h labeled/unlabeled data split and the acoustic model of 100M parameters are used.  }\label{tab:just}
\end{table}

\textbf{Comparing to PL} \  PL is a commonly used approach for semi-supervised learning.  We conduct the comparison with BL-JUST and PL on the Switchboard dataset and report the results in Table \ref{tab:comp}. A model of 45M parameters is first trained using SWB300 and then used to decode SWB2000 to generate pseudo-labels. The generated pseudo-labels and the original ground truth labels are combined to train another model of 74M parameters, which is denoted PL in the table. We also conduct a second pass of PL where we use the trained semi-supervised model to re-decode the unlabeled SWB2000 data and repeat the training procedure again to get an updated model which is denoted PL$^{2}$ in the table. The model configurations and training recipes are the same as those of the supervised baselines in table \ref{tab:swb}. We can see that the first pass PL gives 9.5\% WER and the second pass PL improves the WER to 8.9\%.  They are better than the SWB300 supervised baseline but worse than the SWB2000 supervised baseline. However, BL-JUST with a WER 8.1\% outperforms both PL and PL$^{2}$.

\textbf{Comparing to AO} \  We conduct the comparison with BL-JUST and AO on the Switchboard dataset and report the results in Table \ref{tab:comp}. AO carries out supervised training using SWB300 and unsupervised training using SWB2000 alternately in each epoch. Therefore the training can access both labeled and unlabeled data. The supervised training is used to bias the subsequent unsupervised training toward the distribution of labeled data. To make a fair comparison, AO and BL-JUST use the same model size and configuration. They are also trained under the same learning rate schedule. The difference is that BL-JUST leverages joint training while AO only works on one set of data at a time. We can see that BL-JUST outperforms AO -- 8.2\% vs. 9.3\% for the model with 74M parameters and 8.1\% vs. 8.6\% for the model with 109M parameters.

\begin{table}[tbh]
\centering
\begin{tabular}{ l  c  r  c  c  r } \hline
    model           &    hours      &     size      &        SWB   &    CH    &   Avg       \\
                    &    (L/U)      &               &              &          &             \\ \hline\hline
sup. baseline       &    300/--     &     45M       &        7.4   &   15.0   &   11.2      \\
sup. baseline       &   2000/--     &     74M       &        5.5   &    8.5   &    7.0      \\ \hline
PL                  &   2000/300    &     74M       &        6.3   &   12.6   &    9.5      \\
PL$^{2}$            &   2000/300    &     74M       &        6.1   &   11.8   &    8.9      \\  \hline
AO                  &   2000/300    &     74M       &        6.3   &   12.2   &    9.3      \\
AO                  &   2000/300    &    109M       &        6.1   &   11.2   &    8.6      \\  \hline
BL-JUST             &   2000/300    &     74M       &        5.8   &   10.6   &    8.2      \\
BL-JUST             &   2000/300    &    109M       &        5.8   &   10.5   &    8.1      \\ \hline
\end{tabular}
\caption{NIST Hub5 2000 Switchboard and CallHome test set WERs for supervised RNNT baselines and RNNT acoustic models trained under pseudo-labeling, alternating optimization, and BL-JUST. SWB2000 is used as unlabeled data while SWB300 is used as labeled data. }\label{tab:comp}
\end{table}

\subsection{Analysis of loss and gradient}

\begin{figure}[htb]
\centering
\centerline{\includegraphics[width=7cm, height=5cm]{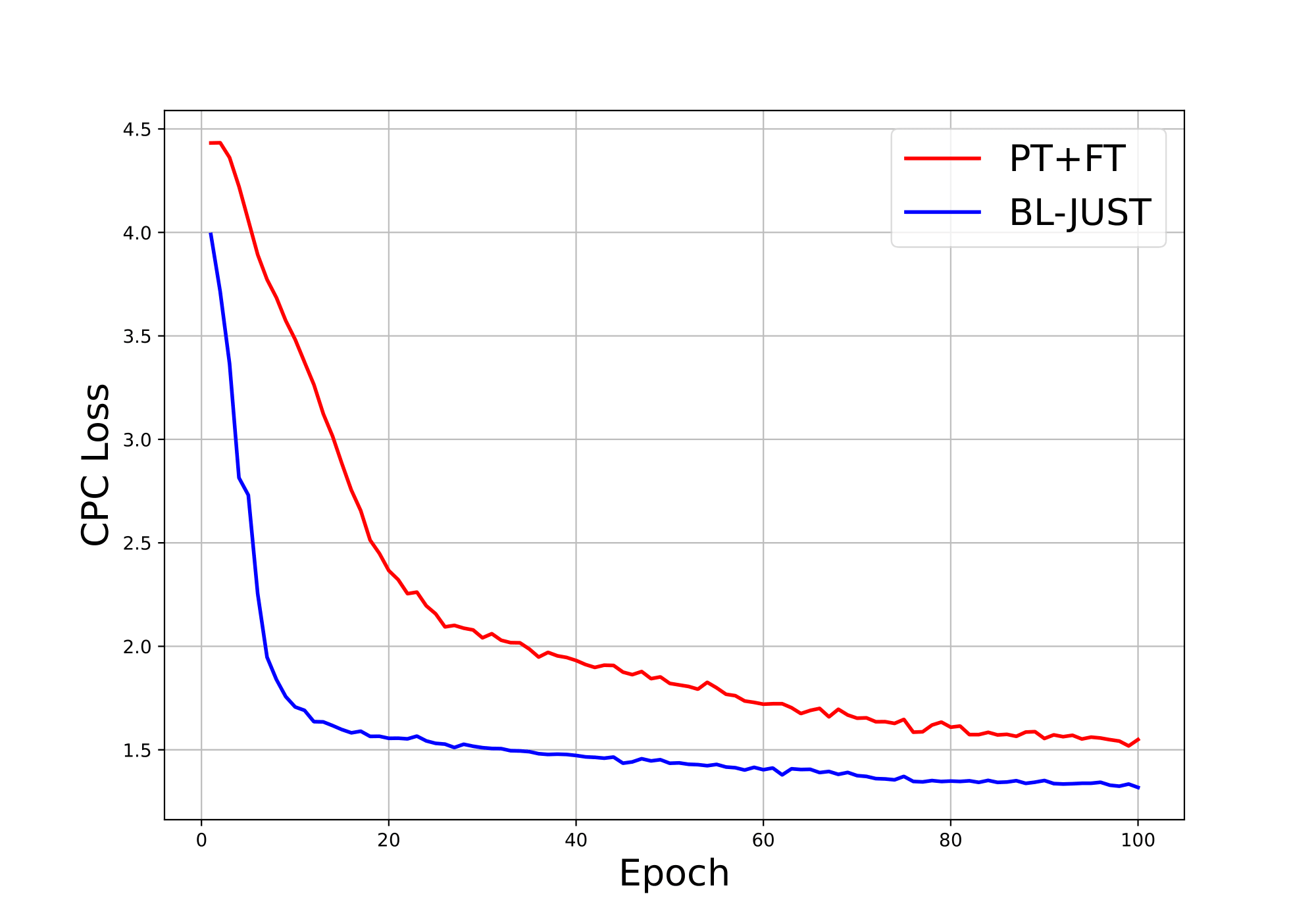}}
\centerline{\includegraphics[width=7cm, height=5cm]{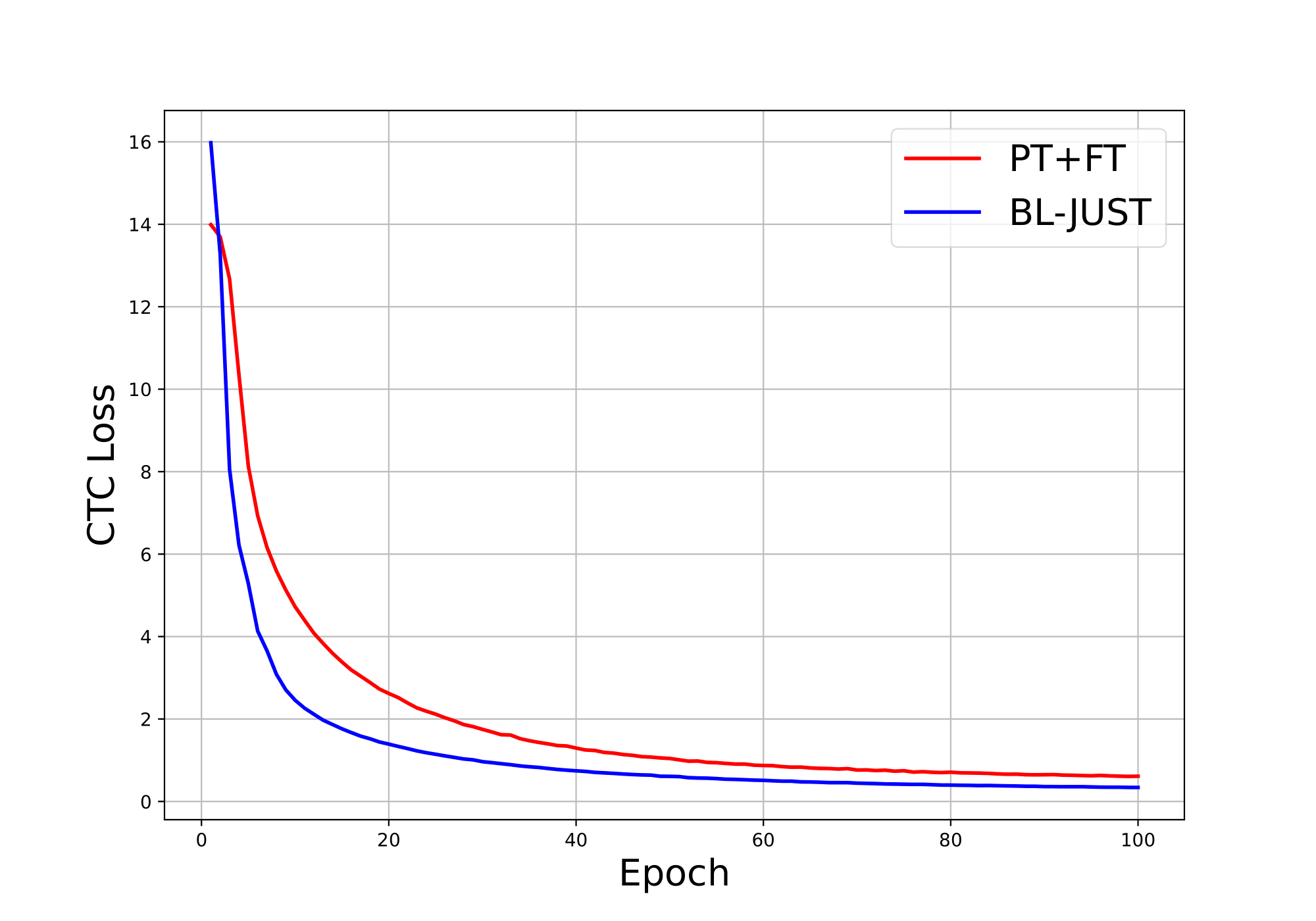}}
\caption{The unsupervised CPC loss (upper panel) and supervised CTC loss (lower panel) of PT+FT and BL-JUST on LibriSpeech using 860 hours of unlabeled data and 100 hours of labeled data (L/U: 100/860).}\label{fig:loss}
\end{figure}

Fig.~\ref{fig:loss} shows the CPC (upper panel) and CTC (lower panel) loss curves when training with PT+FT and BL-JUST on LibriSpeech using 860 hours of unlabeled data and 100 hours of labeled data (L/U: 100/860 in Table~\ref{tab:ls}). In PT+FT, we use 100 epochs of pre-training followed by another 100 epochs of fine-tuning. In BL-JUST, we use 100 epochs of joint training. In BL-JUST training, the upper-level leader problem, which is the CTC supervised loss, provides guidance to the lower-level follower problem, which is the CPC unsupervised loss. Therefore, it can significantly accelerate the reduction of the unsupervised CPC loss compared to PT+FT, as can be clearly observed from Fig.~\ref{fig:loss}. After 100 epochs, BL-JUST yields a lower CPC loss compared to that of PT+FT. Meanwhile, with the constraint from the unsupervised follower problem, BL-JUST also reaches a better CTC loss than PF+FT.  This comparison highlights the effectiveness of BL-JUST in improving CPC and CTC losses simultaneously and its efficiency in quickly reducing the loss.

We also measure the $L_{2}$ norm of the gradient using the models trained under BL-JUST and PT+FT, respectively. They are reported in Table \ref{tab:gradnorm}.  From the table, the norm of the gradient with respect to the supervised CTC loss for both BL-JUST and PT+FT is quite small (0.0041 for BL-JUST and 0.0061 for PT+FT). This shows that both models are close to a local optimum. However, for the gradient with respect to the unsupervised CPC loss, BL-JUST achieves a much smaller norm than that of PT+FT (0.009 for BL-JUST and 0.0245 for PT+FT). This verifies that BL-JUST indeed reaches a shared local optimum of both unsupervised and supervised loss functions.

\begin{table}[tbh]
\centering
\begin{tabular}{ l  c  c } \hline
   model                        &    $||\nabla g||_{2}^{\text{\tiny{CPC}}}$       &     $||\nabla f||_{2}^{\text{\tiny{CTC}}}$    \\ \hline
   PT+FT                        &    0.0245    &    0.0061     \\
   BL-JUST                      &    0.0090     &   0.0041     \\ \hline
\end{tabular}
\caption{The norm of the gradient of using models trained w.r.t. unsupervised CPC loss and supervised CTC loss under PT+FT and BL-JUST on LibriSpeech using 860 hours of unlabeled data and 100 hours of labeled data (L/U: 100/860).}\label{tab:gradnorm}
\end{table}

\section{Summary}
\label{sec:sum}

In this paper, we propose a bilevel training framework, BL-JUST, for ASR acoustic modeling where the upper-level problem is on supervised training loss and the lower-level problem is on unsupervised training loss. The optimal solution of the supervised upper-level problem is constrained by the solution of the unsupervised lower-level problem so that, by solving this bilevel optimization problem, we seek to capture matched (local) optima of both the supervised and unsupervised loss.  It offers the advantage over the widely-used PT+FT strategy where unsupervised and supervised training is disconnected and hence there is no guarantee that the trained model will simultaneously achieve (local) optima of both loss functions.

Extensive experiments are carried out on various datasets (LibriSpeech, Switchboard, and Payload) of various amounts of data (from 100 hours to 90,000 hours) using a variety of unsupervised losses (CPC and BEST-RQ) and supervised losses (CTC and RNNT) and network architectures (Conformer-CTC and Conformer-RNNT). It shows that BL-JUST can consistently outperform PT+FT. Because BL-JUST couples the supervised and unsupervised losses, it can significantly speed up the reduction of unsupervised loss and eventually reach better local optima of both loss functions. We investigated the impact of the penalty factor on the performance of PBGD in BL-JUST and conducted an ablation study on self-supervised exploration and fine-tuning in the implementation. Additionally, we also show that BL-JUST outperforms other existing semi-supervised training strategies such as JUST, pseudo-labeling, and alternating optimization.  All these comparative studies demonstrate the effectiveness and efficiency of the proposed BL-JUST algorithm.

\section{Acknowledgment}
\label{sec:ack}

This work was supported by IBM through the IBM-Rensselaer Future of Computing Research Collaboration.


{\fontsize{9}{9.95}\selectfont
\bibliographystyle{IEEEtran}
\bibliography{ref_tran}

\begin{thebibliography}{10}
\providecommand{\url}[1]{#1}
\csname url@samestyle\endcsname
\providecommand{\newblock}{\relax}
\providecommand{\bibinfo}[2]{#2}
\providecommand{\BIBentrySTDinterwordspacing}{\spaceskip=0pt\relax}
\providecommand{\BIBentryALTinterwordstretchfactor}{4}
\providecommand{\BIBentryALTinterwordspacing}{\spaceskip=\fontdimen2\font plus
\BIBentryALTinterwordstretchfactor\fontdimen3\font minus \fontdimen4\font\relax}
\providecommand{\BIBforeignlanguage}[2]{{%
\expandafter\ifx\csname l@#1\endcsname\relax
\typeout{** WARNING: IEEEtran.bst: No hyphenation pattern has been}%
\typeout{** loaded for the language `#1'. Using the pattern for}%
\typeout{** the default language instead.}%
\else
\language=\csname l@#1\endcsname
\fi
#2}}
\providecommand{\BIBdecl}{\relax}
\BIBdecl

\bibitem{LeCun_DLNature}
Y.~LeCun, Y.~Bengio, and G.~Hinton, ``Deep learning,'' \emph{Nature}, pp. 436--444, May 2015.

\bibitem{Hinton_DNNSPM}
G.~Hinton, L.~Deng, D.~Yu, G.~Dahl, A.~Mohamed, N.~Jaitly, A.~Senior, V.~Vanhoucke, P.~Nguyen, T.~N. Sainath, and B.~Kingsbury, ``Deep neural networks for acoustic modeling in speech recognition,'' \emph{IEEE Signal Processing Maganize}, pp. 82--97, November 2012.

\bibitem{Xiong_ASRParity}
W.~Xiong, J.~Droppo, X.~Huang, F.~Seide, M.~L. Seltzer, A.~Stolcke, D.~Yu, and G.~Zweig, ``Toward human parity in conversational speech recognition,'' \emph{IEEE/ACM Transactions on Audio, Speech, and Language Processing}, vol.~25, no.~12, pp. 2410--2423, 2017.

\bibitem{Saon_2017HumanParity}
G.~Saon, G.~Kurata, T.~Sercu, K.~Audhkhasi, S.~Thomas, D.~Dimitriadis, X.~Cui, B.~Ramabhadran, M.~Picheny, L.-L. Lim, B.~Roomi, and P.~Hall, ``{E}nglish conversational telephone speech recognition by humans and machines,'' in \emph{Interspeech}, 2017, pp. 132--136.

\bibitem{Parthasarathi_MillionhourASR}
S.~H.~K. Parthasarathi and N.~Str\"{o}m, ``Lessons from building acoustic models with a million hours of speech,'' in \emph{International Conference on Acoustics, Speech and Signal Processing (ICASSP)}, 2019, pp. 6670--6674.

\bibitem{Parthasarathi_PetaAM}
S.~H.~K. Parthasarathi, N.~Sivakrishnan, P.~Ladkat, and N.~Strom, ``Realizing petabyte scale acoustic modeling,'' \emph{IEEE Journal on Emerging and Selected Topics in Circuits and Systems}, vol.~9, no.~2, pp. 422--432, 2019.

\bibitem{deng2013new}
L.~Deng, G.~Hinton, and B.~Kingsbury, ``New types of deep neural network learning for speech recognition and related applications: An overview,'' in \emph{IEEE international conference on acoustics, speech and signal processing}, 2013, pp. 8599--8603.

\bibitem{baevski2019vq}
A.~Baevski, S.~Schneider, and M.~Auli, ``vq-wav2vec: Self-supervised learning of discrete speech representations,'' \emph{arXiv preprint arXiv:1910.05453}, 2019.

\bibitem{baevski2020wav2vec}
A.~Baevski, Y.~Zhou, A.~Mohamed, and M.~Auli, ``wav2vec 2.0: A framework for self-supervised learning of speech representations,'' \emph{Advances in neural information processing systems}, vol.~33, pp. 12\,449--12\,460, 2020.

\bibitem{hsu2021hubert}
W.-N. Hsu, B.~Bolte, Y.-H.~H. Tsai, K.~Lakhotia, R.~Salakhutdinov, and A.~Mohamed, ``Hubert: Self-supervised speech representation learning by masked prediction of hidden units,'' \emph{IEEE/ACM Transactions on Audio, Speech, and Language Processing}, vol.~29, pp. 3451--3460, 2021.

\bibitem{Chiu_Bestrq}
C.-C. Chiu, J.~Qin, Y.~Zhang, J.~Yu, and Y.~Wu, ``Self-supervised learning with random-projection quantizer for speech recognition,'' in \emph{International Conference on Machine Learning (ICML)}, 2022, pp. 3915--3924.

\bibitem{rosenstein2005transfer}
M.~T. Rosenstein, Z.~Marx, L.~P. Kaelbling, and T.~G. Dietterich, ``To transfer or not to transfer,'' in \emph{NIPS workshop on transfer learning}, no.~3, 2005.

\bibitem{pan2009survey}
S.~J. Pan and Q.~Yang, ``A survey on transfer learning,'' \emph{IEEE Transactions on knowledge and data engineering}, vol.~22, no.~10, pp. 1345--1359, 2009.

\bibitem{wang2019characterizing}
Z.~Wang, Z.~Dai, B.~P{\'o}czos, and J.~Carbonell, ``Characterizing and avoiding negative transfer,'' in \emph{Proceedings of Conference on Computer Vision and Pattern Recognition}, 2019, pp. 11\,293--11\,302.

\bibitem{Choromanska2015}
A.~Choromanska, M.~Henaff, M.~Mathieu, G.~Ben~Arous, and Y.~LeCun, ``The loss surfaces of multilayer networks,'' in \emph{Proc. International Conference on Artificial Intelligence and Statistics (AISTATS)}, 2015, pp. 192--204.

\bibitem{liu2021investigating}
R.~Liu, J.~Gao, J.~Zhang, D.~Meng, and Z.~Lin, ``Investigating bi-level optimization for learning and vision from a unified perspective: A survey and beyond,'' \emph{IEEE Transactions on Pattern Analysis and Machine Intelligence}, vol.~44, no.~12, pp. 10\,045--10\,067, 2021.

\bibitem{crockett2022bilevel}
C.~Crockett and J.~A. Fessler, ``Bilevel methods for image reconstruction,'' \emph{Foundations and Trends{\textregistered} in Signal Processing}, vol.~15, no. 2-3, pp. 121--289, 2022.

\bibitem{chen2023learning}
L.~Chen, S.~T. Jose, I.~Nikoloska, S.~Park, T.~Chen, and O.~Simeone, ``Learning with limited samples: Meta-learning and applications to communication systems,'' \emph{Foundations and Trends{\textregistered} in Signal Processing}, vol.~17, no.~2, pp. 79--208, 2023.

\bibitem{lu2023meta}
S.~Lu and T.~Gao, ``{Meta-DAG}: Meta causal discovery via bilevel optimization,'' in \emph{IEEE International Conference on Acoustics, Speech and Signal Processing (ICASSP)}, 2023.

\bibitem{franceschi2018bilevel}
L.~Franceschi, P.~Frasconi, S.~Salzo, R.~Grazzi, and M.~Pontil, ``Bilevel programming for hyperparameter optimization and meta-learning,'' in \emph{International Conference on Machine Learning}, 2018, pp. 1568--1577.

\bibitem{finn2017model}
C.~Finn, P.~Abbeel, and S.~Levine, ``Model-agnostic meta-learning for fast adaptation of deep networks,'' in \emph{International Conference on Machine Learning}, 2017, pp. 1126--1135.

\bibitem{shen2023penalty}
H.~Shen and T.~Chen, ``On penalty-based bilevel gradient descent method,'' \emph{arXiv preprint arXiv:2302.05185}, 2023.

\bibitem{bai2022joint}
J.~Bai, B.~Li, Y.~Zhang, A.~Bapna, N.~Siddhartha, K.~C. Sim, and T.~N. Sainath, ``Joint unsupervised and supervised training for multilingual {ASR},'' in \emph{IEEE International Conference on Acoustics, Speech and Signal Processing (ICASSP)}, 2022, pp. 6402--6406.

\bibitem{Xu_IPL}
Q.~Xu, T.~Likhomanenko, J.~Kahn, A.~Hannun, G.~Synnaeve, and R.~Collobert, ``Iterative pseudo-labeling for speech recognition,'' in \emph{Interspeech}, 2020.

\bibitem{Saif_bljust}
A.~Saif, X.~Cui, H.~Shen, S.~Lu, B.~Kingsbury, and T.~Chen, ``Joint unsupervised and supervised training for automatic speech recognition via bilevel optimization,'' in \emph{International Conference on Acoustics, Speech and Signal Processing (ICASSP)}, 2024, pp. 10\,931--10\,935.

\bibitem{sutskever2015towards}
I.~Sutskever, R.~Jozefowicz, K.~Gregor, D.~Rezende, T.~Lillicrap, and O.~Vinyals, ``Towards principled unsupervised learning,'' \emph{arXiv preprint arXiv:1511.06440}, 2015.

\bibitem{Zhang_Bilevel}
Y.~Zhang, P.~Khanduri, I.~Tsaknakis, Y.~Yao, M.~Hong, and S.~Liu, ``An introduction to bilevel optimization: Foundations and applications in signal processing and machine learning,'' \emph{IEEE Signal Processing Maganize}, pp. 38--59, April 2024.

\bibitem{Lu_Bilevel}
Z.~Lu and S.~Mei, ``First-order penalty methods for bilevel optimization,'' \emph{SIAM Journal on Optimization}, vol.~34, no.~2, pp. 1937--1969, 2024.

\bibitem{Kwon_penaltyBL}
J.~Kwon, D.~Kwon, S.~Wright, and R.~D. Nowak, ``On penalty methods for nonconvex bilevel optimization and first-order stochastic approximation,'' in \emph{International Conference on Learning Representations (ICLR)}, 2024.

\bibitem{bottou2018optimization}
L.~Bottou, F.~E. Curtis, and J.~Nocedal, ``Optimization methods for large-scale machine learning,'' \emph{SIAM review}, vol.~60, no.~2, pp. 223--311, 2018.

\bibitem{Zhang_USM}
Y.~Z. et. al., ``Google {USM}: scaling automatic speech recognition beyond 100 languages,'' \emph{arXiv preprint arXiv:2303.01037}, 2023.

\bibitem{panayotov2015librispeech}
V.~Panayotov, G.~Chen, D.~Povey, and S.~Khudanpur, ``Librispeech: an asr corpus based on public domain audio books,'' in \emph{International Conference on Acoustics, Speech and Signal Processing (ICASSP)}, 2015.

\bibitem{Gulati_conformer}
A.~Gulati, J.~Qin, C.-C. Chiu, N.~Parmar, Y.~Zhang, J.~Yu, W.~Han, S.~Wang, Z.~Zhang, Y.~Wu, and R.~Pang, ``Conformer: Convolution-augmented transformer for speech recognition,'' in \emph{Interspeech}, 2020, pp. 5036--5040.

\bibitem{kudo2018sentencepiece}
T.~Kudo and J.~Richardson, ``Sentencepiece: A simple and language independent subword tokenizer and detokenizer for neural text processing,'' \emph{arXiv preprint arXiv:1808.06226}, 2018.

\bibitem{Park_SpecAug}
D.~S. Park, W.~Chan, Y.~Zhang, C.-C. Chiu, B.~Zoph, E.~D. Cubuk, and Q.~V. Le, ``{SpecAugment}: {A} simple data augmentation method for automatic speech recognition,'' in \emph{Interspeech}, 2019, pp. 2613--2617.

\bibitem{oord2018representation}
A.~v.~d. Oord, Y.~Li, and O.~Vinyals, ``Representation learning with contrastive predictive coding,'' \emph{arXiv preprint arXiv:1807.03748}, 2018.

\bibitem{graves2006connectionist}
A.~Graves, S.~Fern{\'a}ndez, F.~Gomez, and J.~Schmidhuber, ``Connectionist temporal classification: labelling unsegmented sequence data with recurrent neural networks,'' in \emph{Proceedings of the 23rd international conference on Machine learning}, 2006, pp. 369--376.

\bibitem{Graves_RNNT}
A.~Graves, ``Sequence transduction with recurrent neural networks,'' \emph{arXiv preprint arXiv:1211.3711}, 2012.

\bibitem{Graves_RNNASR}
{A. Graves and A.-r. Mohamed and G. Hinton}, ``Speech recognition with deep recurrent neural networks,'' in \emph{International Conference on Acoustics, Speech and Signal Processing (ICASSP)}, 2013, pp. 6645--6649.

\bibitem{Saon_RNNT}
G.~Saon, Z.~Tueske, D.~Bolanos, and B.~Kingsbury, ``Advancing {RNN} transducer technology for speech recognition,'' in \emph{International Conference on Acoustics, Speech and Signal Processing (ICASSP)}, 2021.

\bibitem{Li_RNNT}
J.~Li, R.~Zhao, H.~Hu, and Y.~Gong, ``Improving {RNN} transducer modeling for end-to-end speech recognition,'' in \emph{Automatic Speech Recognition and Understanding Workshop (ASRU)}, 2019.

\bibitem{Saon_mixup}
G.~Saon, Z.~Tuske, K.~Audhkhasi, and B.~Kingsbury, ``Sequence noise injected training for end-to-end speech recognition,'' in \emph{International Conference on Acoustics, Speech and Signal Processing (ICASSP)}, 2019, pp. 6261--6265.

\bibitem{Wan_dropconnect}
L.~Wan, M.~Zeiler, S.~Zhang, Y.~LeCun, and R.~Fergus, ``Regularization of neural networks using {DropConnect},'' in \emph{Proceedings of the 35th International Conference on Machine Learning (ICML)}, 2013, pp. 1058--1066.

\bibitem{Saon_RNNTdecoding}
G.~Saon, Z.~Tuske, and K.~Audhkhasi, ``Alignment-length synchronous decoding for {RNN} transducer,'' in \emph{International Conference on Acoustics, Speech and Signal Processing (ICASSP)}, 2020, pp. 7804--7808.

\bibitem{Kingma_adam}
D.~P. Kingma and J.~L. Ba, ``{ADAM}: a method for stochastic optimization,'' in \emph{International Conference on Learning Representations (ICLR)}, 2015.

\bibitem{Smith_OneCycleLR}
L.~N. Smith and N.~Topin, ``Super-convergence: very fast training of neural networks using large learning rates,'' in \emph{Artificial Intelligence and Machine Learning for Multi-Domain Operations Applications}, 2019.

\end{thebibliography}
}


\appendices

\section{Training Recipes for LibriSpeech Experiments}
\label{app:libri}

\textbf{Baseline} In the baseline supervised training, the maximum learning rate is $5e \!-\!4$. A learning rate scheduler is used to monitor the test loss every 10 epochs. If the loss does not decrease during these 10 epochs, the learning rate is reduced by a factor of 0.1. The model is trained for a total of 100 epochs.

\textbf{PT+FT} In the PT+FT training, the unsupervised CPC pre-training starts with a learning rate of $5e\!-\!3$ for 40 epochs, which is then annealed by a factor of 0.1 every 20 epochs. We utilize a context length of 20 frames (200 ms) and predict the next 12 frames, employing 12 negative samples for contrastive loss. The pre-training phase lasts for a total of 100 epochs. Following this, supervised fine-tuning is performed on the pre-trained model with a learning rate of $5e \! \!-4$, using the same learning rate scheduler as in the baseline training. The fine-tuning phase also runs for 100 epochs.

\textbf{BL-JUST} In the BL-JUST training, the training configuration for lower-level unsupervised CPC training is the same as that used in the pre-training of PT+FT. The learning rates are $\alpha = 5e\!-\!3$ and $\beta = 5e \!-\!4$. The penalty factor $\gamma$ starts from 0 and monotonically and linearly increases to 0.2 over epochs with a rate of 0.002. Utilizing the same learning rate scheduler as in the fine-tuning phase of the PT+FT method, BL-JUST training is conducted for 100 epochs. This is followed by an additional fine-tuning phase of 20 epochs with a learning rate of $5e \! \!-5$.

All training uses the AdamW optimizer~\cite{Kingma_adam}. The batch size for supervised training is 128 utterances and 512 utterances for unsupervised training.



\section{Training Recipes for Switchboard Experiments}
\label{app:swb}

\textbf{Baseline} \ In baseline supervised training, the maximum learning rate is $1e\!-\!3$. It starts at $1e\!-\!4$ in the first epoch and then linearly scales up to $1e\!-\!3$ in the first 10 epochs. It holds for another 5 epochs before being annealed by $\frac{1}{\sqrt{2}}$ every epoch afterward. The training ends after 30 epochs.

\textbf{PT+FT} \ In the PT+FT training, unsupervised BEST-RQ pre-training starts training with a learning rate $2e\!-\!4$ for 20 epochs which is then annealed by $\frac{1}{\sqrt{2}}$ every epoch afterward. The pre-training ends after 30 epochs. Then the supervised fine-tuning using a learning rate $2e\!-\!4$ for 5 epochs which is annealed by $\frac{1}{\sqrt{2}}$ every epoch afterward. The fine-tuning ends after 15 epochs. The masking probability in BEST-RQ is 0.02. The mask span is 20 frames. The masked frames are replaced with Gaussian noise with with 0 mean and 0.1 variance. The size of the random codebook is 128.

\textbf{BL-JUST} \ In the BL-JUST training, the configuration of lower-level BEST-RQ is the same as that in the pre-training in PT+FT. The learning rates are $\alpha=2e\!-\!4$ and $\beta=2e\!-\!4$ for first 20 epochs which are then annealed by $\frac{1}{\sqrt{2}}$ every epoch afterwards. The penalty factor $\gamma$ starts from 0 and monotonically and linearly increases to 0.2 over epochs with a rate of 0.007. The final supervised fine-tuning using a learning rate $5e\!-\!5$ for 5 epochs which is annealed by $\frac{1}{\sqrt{2}}$ every epoch afterward. The fine-tuning ends after 15 epochs.

All training uses the AdamW optimizer~\cite{Kingma_adam}. The batch size for supervised training is 128 utterances and 512 utterances for unsupervised training.

\section{Training Recipes for Payload experiments}
\label{app:payload}

\textbf{Baseline} \ The supervised baseline is trained using the OneCycleLR~\cite{Smith_OneCycleLR} training schedule. The maximum learning rate is $5e\!-\!4$ and it starts with a linear warmup phase from $5e\!-\!5$ to $5e\!-\!4$ over
the first 9 epochs followed by a linear annealing phase to 0 for the next 11 epochs. The training takes 20 epochs. The conformer has 10 conformer blocks. This is a highly optimized high-performing baseline.

\textbf{PT+FT} \ In the PT+FT training, unsupervised BEST-RQ pre-training starts training with a learning rate $2e\!-\!4$ for 5 epochs which is then annealed by $\frac{1}{\sqrt{2}}$ every epoch afterward. The pre-training ends after 10 epochs. Then the supervised fine-tuning using a learning rate $2e\!-\!4$ for 5 epochs which is annealed by $\frac{1}{\sqrt{2}}$ every epoch afterward. The fine-tuning ends after 15 epochs. The masking probability in BEST-RQ is 0.02. The mask span is 20 frames. The masked frames are replaced with Gaussian noise with with 0 mean and 0.1 variance. The size of the random codebook is 256. The conformer has 12 conformer blocks.

\textbf{BL-JUST} \  In the BL-JUST training, the configuration of BEST-RQ is the same as that in the pre-training in PT+FT. The learning rates are $\alpha=2e\!-\!4$ and $\beta=2e\!-\!4$ for first 5 epochs which are then annealed by $\frac{1}{\sqrt{2}}$ every epoch afterwards. The penalty factor $\gamma$ starts from 0 and monotonically and linearly increases to 0.1 over epochs. The final supervised fine-tuning using a learning rate $5e\!-\!5$ for 5 epochs which is annealed by $\frac{1}{\sqrt{2}}$ every epoch afterward. The fine-tuning ends after 15 epochs. The conformer has 12 conformer blocks.

All training uses the AdamW optimizer.  The batch size for supervised training is 256 utterances and 512 utterances for unsupervised training.

\end{document}